\documentclass[11pt]{article}

\usepackage[preprint]{acl}

\usepackage{times}
\usepackage{latexsym}
\usepackage[T1]{fontenc}
\usepackage[utf8]{inputenc}
\usepackage{microtype}
\usepackage{inconsolata}

\usepackage{amsmath,amssymb}
\usepackage{graphicx}
\usepackage{booktabs}
\usepackage{amsfonts}
\usepackage{nicefrac}
\usepackage{multirow}
\usepackage{enumitem}
\usepackage{cleveref}
\usepackage{xspace}
\usepackage{algorithm}
\usepackage{algorithmic}
\usepackage{subcaption}
\usepackage{verbatim}
\usepackage{tabularx}
\usepackage{array}
\usepackage{placeins}
\usepackage[most]{tcolorbox}
\usepackage{tikz}
\usetikzlibrary{arrows.meta,positioning}
\definecolor{PromptBlue}{HTML}{F1F6FE}
\definecolor{PromptGreen}{HTML}{F1FAF6}
\definecolor{PromptOrange}{HTML}{FFF7ED}
\definecolor{PromptGray}{HTML}{FBFCFE}
\definecolor{PromptFrame}{HTML}{C7D0DB}
\tcbset{
  samplebox/.style={
    enhanced,
    boxrule=0.35pt,
    arc=1.5pt,
    left=4pt,
    right=4pt,
    top=3pt,
    bottom=3pt,
    colframe=PromptFrame,
    fonttitle=\bfseries\scriptsize,
    coltitle=black,
    colbacktitle=white,
    attach boxed title to top left={xshift=4pt,yshift=-1.3mm},
    boxed title style={boxrule=0.2pt,arc=1pt,left=2pt,right=2pt,top=1pt,bottom=1pt,colframe=PromptFrame,colback=white},
  }
}

\newcommand{\method}{SymPlan\xspace}
\newcommand{\eg}{\textit{e.g.}\xspace}

\newcommand{\etal}{\textit{et al.}\xspace}

\title{Plan Right, Then Plan Tight:\\
Symbolic RL for Efficient Embodied Reasoning}

\author{%
  Xiangli Shi$^{1}$\quad
  Xiaomeng Zhu$^{2}$\quad
  Ye Tian$^{3}$\quad
  Yuchun Guo$^{3}$\quad
  Ziyang Sun$^{4}$ \\
  \bfseries
  Lujie Yin$^{1}$\quad
  Yuxuan Zhou$^{1}$\quad
  Yufei Huang$^{3}$\thanks{~Corresponding author.} \\
  \vspace{2pt} \\
  {\normalfont $^{1}$Tsinghua University \quad
  $^{2}$The Hong Kong University of Science and Technology} \\
  {\normalfont $^{3}$Tencent \quad
  $^{4}$University of London} \\
  \vspace{2pt} \\
  {\normalfont \texttt{shixl25@mails.tsinghua.edu.cn}} \\
}

\begin{document}
\maketitle

\begin{abstract}
Embodied task planning asks an agent to turn a natural-language
instruction into an executable sequence of actions in a physical
scene, and is a building block for household, assistive, and
service robots. Recent prompting-based and reinforcement-learning
planners generate fluent action text but lack a cheap deterministic
check that the produced plan is valid in the target world, while
high-fidelity simulation is too slow to serve as an inner-loop
training signal. The general problem is therefore how to obtain
verifiable supervision and rewards for embodied planners without
relying on string-level matching or full simulation. Here we show
that a single BDDL specification, automatically constructed from
open-world video evidence or curated tasks, can serve as a shared
interface for data construction, plan verification, and reward
design. A video-to-BDDL parser, an LLM verifier, and a lightweight
symbolic engine together supply dense feedback at millisecond
latency. We further introduce GroupAdapt, a difficulty-aware length
schedule that uses the in-batch group pass rate as a zero-cost
signal so that hard prompts get wider length tolerance and
automatically tighten as their pass rate improves. Under the guidance
of the proposed verifier and GroupAdapt schedule, the 8B planner
attains a Strict-Pass score of 97.3 on BEHAVIOR-1000, yielding a
25.9\% relative improvement over the Qwen3-8B baseline. This result
exceeds the strongest large-model baseline by 3.5\%, while
simultaneously compressing the response length by 79\% to 207 tokens,
demonstrating both effectiveness and efficiency.
\end{abstract}

\begin{figure}[t]
\centering
\includegraphics[width=\linewidth]{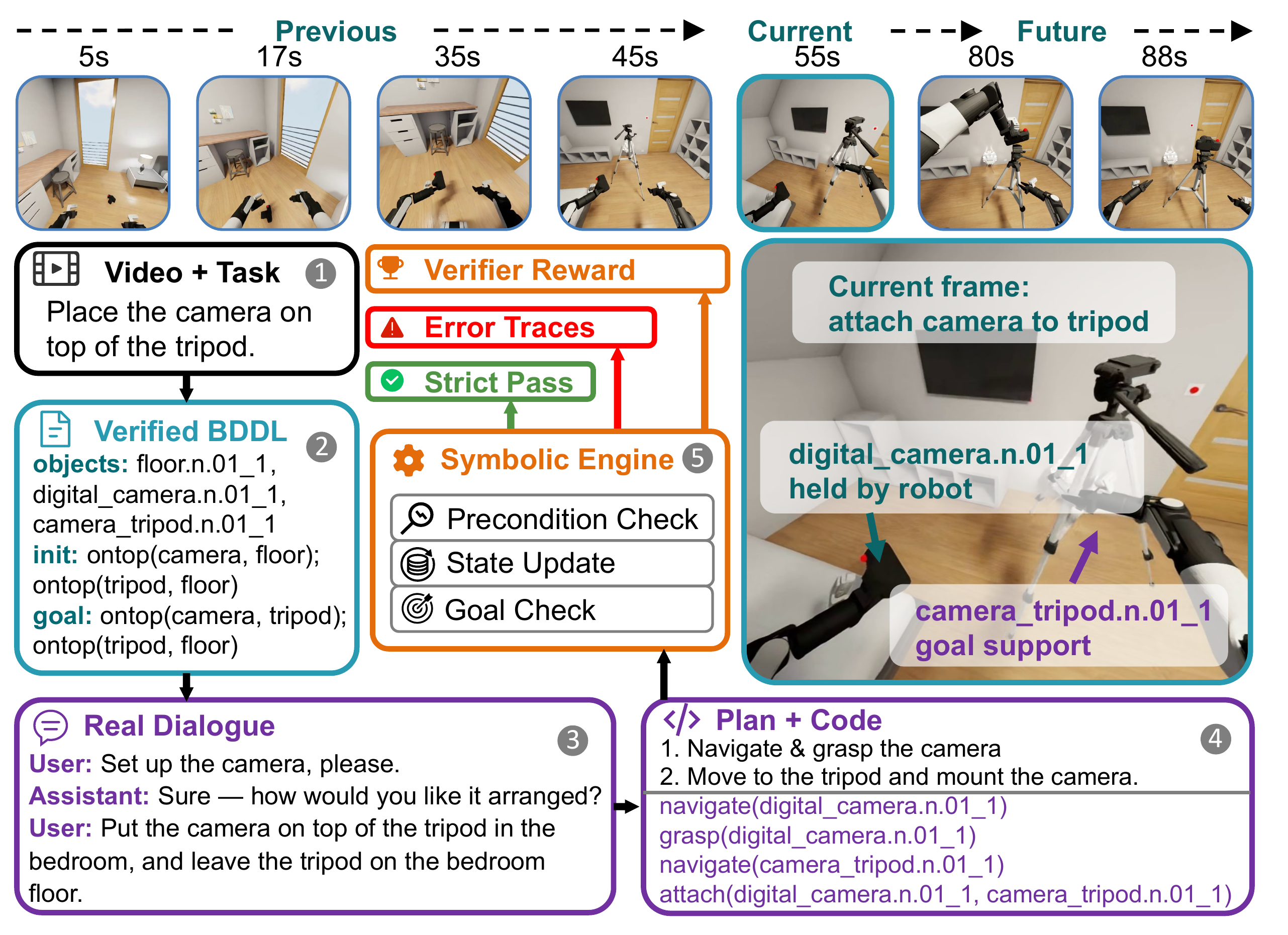}
\caption{\textbf{A running example of \method based on a real
\texttt{attach\_a\_camera\_to\_a\_tripod} guided sample.} Open-world
video evidence is parsed into grounded objects and initial
predicates, verified into BDDL, rewritten as planner-facing
dialogue, turned into executable action code, and checked by the
symbolic engine. The same BDDL interface therefore builds data,
verifies plans, and supplies training rewards.}
\label{fig:intro_teaser}
\end{figure}

\section{Introduction}
\label{sec:intro}

Embodied task planning studies how an agent should transform a
high-level instruction and its surrounding world state into an
executable sequence of actions~\cite{ahn2022can,singh2022progprompt}. This
problem is increasingly important as
robots move from scripted demonstrations toward open-ended household
and assistive settings: a command such as ``attach the camera to the
tripod'' is not solved by emitting a verb alone, but by grounding
the intended digital camera and tripod in the scene, recognizing
that both are initially on the floor, selecting executable
\texttt{grasp} and \texttt{attach} actions, and reaching a goal
state \texttt{(attached digital\_camera camera\_tripod)} that can
be verified. A practical robot stack therefore organizes itself in
layers: a low-level VLA or skill controller turns short subgoals
into motor actions~\cite{zitkovich2023rt}, while a
high-level planner decides which subgoals to issue, in what order,
and against which world state. Our work targets the
\emph{planning-brain} layer in between, which converts a
natural-language instruction and the current scene into a verified
symbolic task structure that the lower layer can execute.
Existing solutions to this layer are split: classical PDDL-style and
neuro-symbolic planners emphasize formal executability under
hand-specified domains~\cite{fox2003pddl2,ahn2026towards},
while LLM-based planners~\cite{huang2022language,yao2022react} are
fluent on open-world tasks but lack a cheap,
deterministic check that the produced plan is valid in the target
world.

A natural interface for closing this gap is BDDL, the Behavior
Domain Definition Language used in BEHAVIOR-1K and related
benchmarks~\cite{li2023behavior}. A BDDL specification encodes
the typed objects in a scene, predicates describing the initial
state, and predicates defining the goal condition. Although BDDL
shares first-order logic with classical
PDDL~\cite{fox2003pddl2}, it differs in two ways
that matter for our setting: object types are grounded in WordNet
synsets such as \texttt{digital\_camera.n.01}, which makes the
representation open-vocabulary, and BDDL specifies only the scene
together with initial and goal predicates, leaving the action
model to a simulator-backed library rather than requiring every
action schema to be hand-authored. Unlike free-form
language instructions, this structure is directly checkable, and
unlike low-level robot trajectories, it abstracts away
embodiment-specific control and exposes the planning problem
itself. BDDL is therefore a natural interface between open-world
perception and verifiable plan execution.

We therefore study \emph{BDDL-centric embodied planning}: given a
task description and either an open-world video or an existing task
specification, can we construct a verified BDDL representation and
use it to train a compact planning model? The goal is not to
replace VLA control or physics simulation, but to expose a scalable
layer at which task structure can simultaneously provide
supervision, execution checks, and reward signals. In this
formulation, BDDL is not only an offline evaluation artifact. It is
the shared representation that connects data construction, symbolic
execution, RL reward design, and response-length control.
\Cref{fig:intro_teaser} illustrates this setting with a concrete
BDDL-to-execution example.

However, the broader adoption of BDDL still faces several challenges.
First, data construction must recover typed objects, initial
predicates, and goals from noisy sources, whereas existing BDDL files
are usually curated as benchmark metadata~\cite{li2023behavior}.
Second, verification must be strict enough to catch physical and
symbolic errors but fast enough to score many training rollouts,
unlike full simulation loops~\cite{savva2019habitat}. Third, compact
planning must shorten responses only after executable behavior is
reliable. Reasoning models often solve tasks with long
outputs~\cite{yang2025qwen3}, but length pressure applied too early can
remove useful planning steps and reduce correctness~\cite{yuan2025efficient}.
These challenges are coupled because the same representation must
support data construction, plan execution, reward computation, and
compression.

Therefore, we propose \method, a three-stage pipeline centered on
verified BDDL to address these challenges. First, to construct
reliable BDDL specifications from complex visual tasks, \method
employs a video-to-BDDL parser to ground task-relevant objects, infer
initial predicates, and derive goal predicates from the task
description. The resulting draft is further examined by an LLM
verifier, which checks syntax, object consistency, predicate
feasibility, and goal executability before acceptance. Second, to make
BDDL executable and informative for planning-model evaluation,
\method rewrites the verified specification into hierarchical planning
conversations and executes them with a fast symbolic engine. This
engine reports Goal-Completion, Engine-Pass, Strict-Pass, and
diagnostic error traces, while the action library is expanded when new
predicate requirements arise. Third, to translate symbolic
verification into effective model optimization, the same engine is
used to filter SFT data and provide rewards for an SFT-initialized
DAPO policy~\cite{yu2026dapo}. We further incorporate Short-RL's
correctness-gated length penalty and introduce \textbf{GroupAdapt},
which uses the per-prompt group pass rate as a zero-cost difficulty
signal to adapt the length tolerance for easy and hard prompts.

Taken together, \method is organized around three design choices that
are evaluated throughout the paper:
\begin{enumerate}[leftmargin=*]
    \item \textbf{BDDL as the data interface.} We formulate a pipeline
    that converts open-world task evidence and curated task files into
    verified BDDL, then into planner-facing supervision through
    extended-BDDL normalization, hierarchical conversations, and
    automated action discovery
    (\Cref{sec:data_synthesis,app:data,app:action_builder}).
    \item \textbf{Symbolic verification as training feedback.} We build
    a millisecond-latency symbolic engine that verifies generated plans
    against typed objects, initial states, and goal conditions, exposing
    deterministic signals for evaluation, rejection sampling, and reward
    design
    (\Cref{sec:symbolic_engine,sec:reward_design}).
    \item \textbf{Compact executable planning.} We combine
    verifier-driven SFT, symbolic-reward DAPO, and correctness-gated
    length control so that the 8B planner preserves strict-pass
    executability while reducing response length on B-100 and B-1000
    (\Cref{sec:reward_design,sec:length_compression,app:pass_at_k}).
\end{enumerate}

\section{Related Work}
\label{sec:related}

\paragraph{LLM-based task planning.}
Recent work uses LLMs to generate or rank executable plans for
embodied agents, either by prompting pretrained models
directly~\cite{huang2022language,yao2022react,liu2023llm+,singh2022progprompt,song2023llm}, grounding proposals with
affordances or scene structure~\cite{ahn2022can,rana2023sayplan},
or integrating planning into larger embodied systems~\cite{liang2023code,wang2023voyager,driess2023palm,zitkovich2023rt}. These
methods show that language models can express
long-horizon intent, but most rely on prompting, heuristic grounding,
or downstream simulation feedback. We instead train the planner with
deterministic symbolic feedback derived from the task specification.

\paragraph{Symbolic and simulation-based verification.}
Embodied benchmarks such as VirtualHome~\cite{puig2018virtualhome},
AI2-THOR~\cite{kolve2017ai2}, Habitat~\cite{savva2019habitat},
ALFRED~\cite{shridhar2020alfred}, TEACh~\cite{padmakumar2022teach},
and BEHAVIOR-1K / OmniGibson~\cite{li2023behavior} provide rich
household tasks and simulation environments. Classical planning also
offers symbolic representations such as PDDL and STRIPS for plan
checking~\cite{fikes1971strips,fox2003pddl2}, and task-and-motion
planning connects symbolic goals to continuous feasibility~\cite{garrett2021integrated,garrett2020pddlstream}. Recent neurosymbolic work combines LLM
generation with symbolic verification~\cite{ahn2026towards}. Full
simulation is valuable but too expensive for inner-loop RL, while
hand-crafted symbolic domains can be hard to scale. Our verifier is
built directly on BDDL, so it inherits typed objects, predicates, and
goals from the benchmark specification and can execute candidate plans
at millisecond latency.

\paragraph{RL and reasoning compression.}
Group-based RL methods such as GRPO and DAPO have become effective for
training LLM reasoning policies~\cite{shao2024deepseekmath,guo2025deepseek,yu2026dapo}, building on broader work on chain-of-thought
and verifier-guided reasoning~\cite{wei2022chain,kojima2022large,lightman2024let}. Related work also applies RL to embodied or
reflective reasoning~\cite{kim2026robot,shen2025satori}. A separate line
of work reduces verbose reasoning through long-to-short distillation,
length penalties, or difficulty-aware
budgets~\cite{team2025kimi,yuan2025efficient,aggarwal2025l1,liang2025deepcompress}. These methods typically evaluate answers with
string matching or language-level judges. In our setting, the final
reasoning must produce a syntactically valid, verifier-accepting action
sequence, so we gate length compression on symbolic correctness and use
the verifier itself as the reward source.

\section{Method}
\label{sec:method}

The central idea behind \method is that \emph{verified BDDL can serve
as the common interface between open-world data, executable
verification, and RL training}. When a curated BDDL task is available,
its typed objects, initial predicates, and goal predicates can be used
directly. When only a task description and video are available, a
video-to-BDDL parser first grounds task-relevant objects and key
frames, infers initial and goal predicates, and passes a draft
specification to an LLM verifier for checking. Once verified, the
same formal structure can be used to synthesize planner data, execute
candidate plans, and compute training rewards. Because new tasks
enter the pipeline as BDDL drafts rather than hand-authored domains,
\method scales naturally with the task pool: open-world videos and
curated benchmarks share a single interface, and the action library
itself grows on demand as new predicates are observed
(\Cref{sec:data_synthesis,app:action_scaling}).

\begin{figure*}[t]
    \centering
    \includegraphics[width=\linewidth]{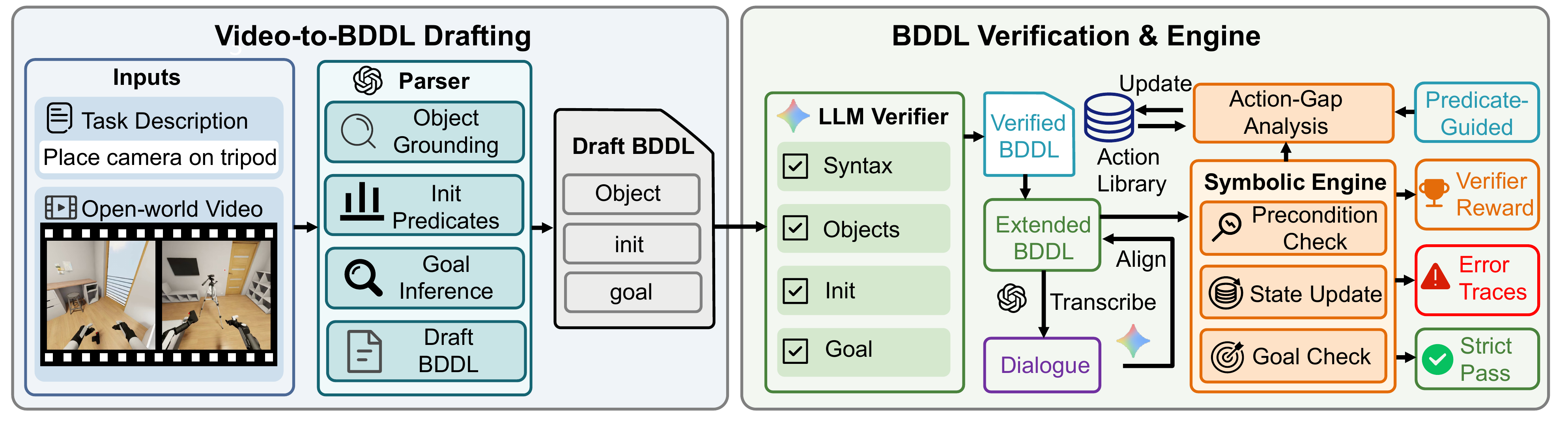}
    \caption{\textbf{BDDL-centric data construction and symbolic verification.}
    A task description and open-world video are parsed into draft BDDL,
    checked by an LLM verifier, and executed by a fast symbolic engine.
    The engine uses the action library, performs predicate-guided action
    expansion when gaps are detected, and returns verifier signals,
    error traces, and verified trajectories.}
    \label{fig:architecture_bddl}
\end{figure*}

\Cref{fig:architecture_bddl} summarizes the BDDL-centric data
construction and verification workflow, which converts task
descriptions and open-world videos into verified BDDL, an expanded
action library, verifier signals, error traces, and verified
trajectories (\Cref{sec:data_synthesis,sec:symbolic_engine}).

\subsection{Video/BDDL-to-Data Synthesis and Action Discovery}
\label{sec:data_synthesis}

\method converts curated BDDL or open-world videos into
verifier-aligned planner supervision. For videos, a parser grounds
task-relevant objects and key frames, infers initial and goal
predicates, and drafts a BDDL specification, and an LLM verifier checks
syntax, object consistency, and goal executability, and only accepted
drafts enter the data factory. Each verified task is normalized to
extended BDDL, a planner-facing expansion of BDDL that preserves the
task-relevant objects, initial predicates, and goals while adding
realistic scene distractors and their predicates to better recover a
cluttered real environment. It is then rewritten into a sample with
environment state, robot embodiment, short clarification dialogue,
and target steps plus executable code (\Cref{app:data}).

When new goal predicates are not covered by the current action set,
an automatic action builder proposes and compiles the missing
callable actions (\Cref{app:action_builder}).

Cold-start SFT uses symbolic rejection sampling with
\textbf{Gemma-4-31B-IT} as the teacher: for each prompt we draw $K$
rollouts, keep the shortest strict-pass trajectory when any sample
passes, and otherwise exclude the prompt. This yields the compact
checkpoint \textbf{D-SFT} that initializes RL.

\subsection{Symbolic Verification Engine}
\label{sec:symbolic_engine}

Given a generated plan $y$ consisting of an action sequence
$\langle a_1, a_2, \ldots, a_n \rangle$, the symbolic engine
executes each action against a state representation derived from
the BDDL task specification~\cite{li2023behavior}.

\paragraph{State representation.}
Each task defines an initial state $s_0$ (a set of predicates
over objects, \eg \texttt{(ontop candle table)}) and a goal
state $g$ (a conjunction of target predicates). The engine
maintains a state $s_t$ that is updated by each action's effects.
For example, in the gift-basket task, a valid plan may first open a
basket, then pick a candle from the table, and finally place it in
the basket. If the model skips the open action and directly places
an object into a closed basket, the goal may still become partially
satisfied, but the violated precondition is recorded in the error
set. This is the difference between matching the final goal and
producing an executable plan.

\paragraph{Execution and verification.}
For each action $a_t$, the engine checks whether its
preconditions are satisfied in $s_t$. If satisfied, the state
is updated according to the action's predefined effects
(\texttt{set}/\texttt{clear} specifications). If violated, the
action is logged as an error but execution continues to provide
maximum feedback. After all actions execute, the engine computes:

\begin{itemize}[leftmargin=*,nosep]
  \item \textbf{Goal Completion Rate (GCR):}
    $\mathrm{GCR} = |\{g_i \in g : g_i \text{ satisfied}\}| / |g|$
  \item \textbf{Error set} $\mathcal{E}$: all precondition
    violations encountered during execution.
  \item \textbf{Engine Pass (EP):} $\mathrm{GCR} = 1.0$
    (all goals met, errors permitted).
  \item \textbf{Strict Pass (SP):} $\mathrm{EP} \wedge
    |\mathcal{E}| = 0$ (all goals met, no errors).
\end{itemize}

\subsection{SFT-Initialized RL with Multi-Granular Symbolic Rewards}
\label{sec:reward_design}

Training proceeds in two stages. The cold-start rejection-sampling
procedure of \Cref{sec:data_synthesis} produces \textbf{D-SFT}, which
is the sole initialization for RL. The same symbolic verifier that
labels supervised trajectories also supplies the RL reward.

\paragraph{Optimizer and symbolic reward.}
The reward must distinguish three cases that a binary success signal
would conflate: a plan that reaches only part of the goal, a plan
that reaches the goal after illegal actions, and a strict-pass plan
that is both goal-complete and error-free. We therefore shape the
answer reward with three symbolic signals from the engine: progress
(GCR), legality (precondition errors), and goal completion (EP). We
adopt DAPO~\cite{yu2026dapo}, an extension of
GRPO~\cite{shao2024deepseekmath} with dynamic sampling, asymmetric
clipping (Clip-Higher, $\epsilon_{\text{low}}{=}0.2$,
$\epsilon_{\text{high}}{=}0.28$), token-level policy-gradient
aggregation, and overlong reward shaping. KL regularization is
disabled since the SFT initialization already acts as a strong
prior. The total reward is
$R(y)=R_{\text{fmt}}(y)+R_{\text{ans}}(y)+R_{\text{len}}(y)$,
where $R_{\text{fmt}}$ is a binary tag penalty and $R_{\text{len}}$
is the gated length term of \Cref{sec:length_compression}. The
answer reward in \Cref{eq:answer_reward} combines a
\emph{continuous} base $(-0.5+2.5\,\mathrm{GCR})$, a hard error
penalty of $-1$ when any precondition is violated, and a $+0.5$
engine-pass bonus:
\begin{equation}
  R_{\text{ans}} =
    (-0.5 + 2.5\,\mathrm{GCR})
    - \mathbf{1}_{|\mathcal{E}|>0}
    + 0.5\cdot\mathbf{1}_{\text{EP}}.
  \label{eq:answer_reward}
\end{equation}
For example, a partial plan receives graded credit through GCR, an
engine-pass-only plan loses one point for the illegal action, and a
strict-pass plan receives both full goal credit and the EP bonus.
The continuous base retains variance on hard prompts that a
binary \texttt{is\_successful} signal would filter out via DAPO's
dynamic sampling, while the $\geq\!1$ gap between strict-pass and
engine-pass-only outcomes prevents a ``conservatism trap'' in
which a model trades goal coverage for cleanliness. Full
reward-landscape analysis is in \Cref{app:reward_landscape}.

In the experimental section we evaluate three SFT-initialized
variants under a unified naming scheme: \textbf{DAPO} (plain
symbolic-reward baseline), \textbf{DAPO + Short-RL} (adds the
lazy length penalty), and \textbf{DAPO + Short-RL + GroupAdapt}
(our full method, see \Cref{sec:length_compression}).

\subsection{Correctness-Gated Length Compression}
\label{sec:length_compression}

Following Short-RL~\cite{yuan2025efficient}, response length is
regularized only after the model reaches stable correctness.
GroupAdapt keeps this correctness gate but replaces the fixed
length tolerance with a task-adaptive one. Its key signal is the
\emph{current group pass rate}: for each prompt, DAPO already
samples a group of rollouts, and the fraction of strict-pass
rollouts directly reflects how well the current policy handles that
prompt. This signal is zero-cost because it is produced by the RL
batch itself, requiring no extra difficulty classifier, simulator
call, or manual task label.

\begin{figure*}[!t]
    \centering
    \includegraphics[width=\linewidth]{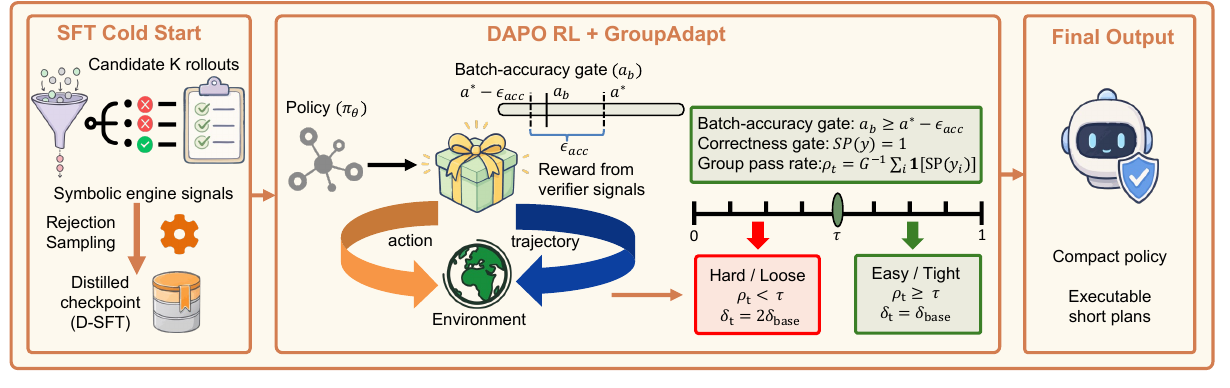}
    \caption{\textbf{SFT-initialized verifiable RL for compact planning.}
    Artifacts from \Cref{fig:architecture_bddl} initialize D-SFT and
    provide symbolic engine signals for DAPO. The batch-accuracy gate
    applies the length penalty only when the current batch accuracy
    exceeds the running maximum minus tolerance, and the penalty is
    charged only to strict-pass samples, and otherwise training continues
    with symbolic correctness rewards. GroupAdapt uses the current
    group pass rate to assign an easy budget $\delta_{\text{base}}$ or
    a hard budget $2\delta_{\text{base}}$, so hard prompts first
    receive wider tolerance and automatically switch to the easy budget
    as their pass rate improves.}
    \label{fig:architecture_rl}
\end{figure*}

\paragraph{Accuracy gate.}
Let $a_b$ be the current batch SP rate and $a^*$ its running best.
The length reward is gated: $R_{\text{len}}=0$ whenever
$\neg\,\text{SP}(y)$ or $a_b<a^*-\epsilon_{\text{acc}}$. Hence the
model first learns \emph{what} to plan before being pushed to plan
\emph{concisely}.

\paragraph{Group-adaptive tolerance.}
For each prompt $t$ with group size $G{=}8$, let
$\rho_t = G^{-1}\sum_{i=1}^{G}\mathbf{1}[\mathrm{SP}(y_i)]$
be the instantaneous strict-pass rate in the current DAPO batch. We set
$\delta_{\text{base}}{=}200$ tokens as the easy-task budget and use a
binary easy/hard schedule with threshold $\tau{=}0.5$:
\begin{equation}
  \delta_t =
  \begin{cases}
    \delta_{\text{base}} & \rho_t \ge \tau \quad (\text{easy})\\
    2\,\delta_{\text{base}} & \rho_t < \tau \quad (\text{hard})
  \end{cases}
  \label{eq:adaptive_tolerance}
\end{equation}
Short-RL uses a fixed $200$-token budget for every prompt. GroupAdapt
keeps this $200$-token tolerance for easy prompts, but gives hard
prompts $400$ tokens of slack. Thus difficult tasks first receive a
wider length tolerance, and as their group pass rate improves beyond
$\tau$, they are automatically reassigned to the easy budget.

\paragraph{Length reward.}
When the accuracy gate is open, $R_{\text{len}}$ in
\Cref{eq:len_reward} rewards being within $\delta_t$ tokens of the
shortest known SP length $\ell^{\min}_t$ for task $t$ and otherwise
decays linearly up to the observed range
$\Delta_t=\ell^{\max}_t-\ell^{\min}_t+\varepsilon$:
\begin{equation}
  R_{\text{len}}=
  \begin{cases}
    0.5 & \ell(y)\le\ell^{\min}_t+\delta_t\\[2pt]
    0.5-\dfrac{\ell(y)-\ell^{\min}_t}{\Delta_t} & \text{otherwise.}
  \end{cases}
  \label{eq:len_reward}
\end{equation}
Per-task references $(\ell^{\min}_t,\ell^{\max}_t)$ are updated
online using only SP responses, so the compression target always
reflects achievable quality. Early in training the gate is closed
and the answer reward drives correctness, and once SP stabilizes, the
gate opens and GroupAdapt matches Short-RL on easy tasks while
protecting hard tasks with a wider $400$-token tolerance. When a
hard prompt's group pass rate improves, it automatically moves back
to the easy budget. \Cref{fig:architecture_rl} summarizes this
SFT-initialized RL workflow.

\section{Experiments}
\label{sec:experiments}

\begin{table*}[!t]
\centering
\caption{One-sample symbolic evaluation on BEHAVIOR-1K under the
Guided extended-BDDL prompt. SP\% = Strict-Pass rate, EP\% =
Engine-Pass rate, GCR = average Goal Completion Ratio (\%), and
Err\% = engine error rate among valid samples. DAPO-stage variants are
initialized from D-SFT and averaged over three decoding seeds.}
\label{tab:main_results}
\footnotesize
\setlength{\tabcolsep}{4pt}
\begin{tabular}{@{}llcccccccc@{}}
\toprule
& & \multicolumn{4}{c}{\textbf{B-1000}} &
  \multicolumn{4}{c}{\textbf{B-100}} \\
\cmidrule(lr){3-6} \cmidrule(lr){7-10}
\textbf{Model} & \textbf{Size} &
  SP$\uparrow$ & EP$\uparrow$ & GCR$\uparrow$ & Err$\downarrow$ &
  SP$\uparrow$ & EP$\uparrow$ & GCR$\uparrow$ & Err$\downarrow$ \\
\midrule
\multicolumn{10}{l}{\textit{Frontier API / large open models}} \\
DeepSeek-V4-Flash        & 284B & 93.8 & 96.9 & 98.36 & 3.6 & 90.0 & 96.0 & 98.74 & 6.5 \\
DeepSeek-V4-Pro          & 1.6T & 89.7 & 94.8 & 96.92 & 6.2 & 86.5 & 95.5 & 98.24 & 9.5 \\
Gemini-3.1-Pro           & --   & 87.1 & 87.6 & 89.39 & 0.5 & 97.5 & 97.5 & 99.19 & 0.0 \\
Kimi-K2.6                & 1T   & 92.3 & 95.9 & 96.88 & 4.6 & 93.0 & 97.5 & 99.36 & 4.5 \\
GLM-5.1                  & 754B & 90.7 & 94.3 & 96.88 & 6.2 & 93.0 & 96.5 & 98.40 & 5.5 \\
GPT-5.4                  & --   & 92.8 & 93.8 & 95.34 & 1.0 & 97.0 & 97.0 & 99.20 & 1.0 \\
Qwen3.5-122B-A10B        & 122B & 86.6 & 90.7 & 93.80 & 7.2 & 82.5 & 93.0 & 96.93 & 13.5 \\
\midrule
\multicolumn{10}{l}{\textit{Small / medium open models}} \\
Gemma-4-31B-IT           & 31B  & 92.8 & 95.4 & 96.49 & 4.6 & 94.5 & 96.5 & 99.21 & 2.0 \\
Qwen3.6-35B-A3B          & 35B  & 88.1 & 93.3 & 95.17 & 9.3 & 83.0 & 94.0 & 96.62 & 15.5 \\
Qwen3.5-35B-A3B          & 35B  & 76.3 & 90.7 & 93.14 & 19.1 & 69.0 & 82.5 & 86.81 & 23.5 \\
Qwen3.6-27B              & 27B  & 91.2 & 94.8 & 97.21 & 4.1 & 90.5 & 96.5 & 98.49 & 6.5 \\
Qwen3.5-27B              & 27B  & 83.0 & 89.7 & 93.71 & 12.4 & 80.0 & 92.5 & 94.91 & 18.5 \\
Qwen3-8B                 & 8B   & 77.3 & 83.0 & 89.84 & 13.4 & 72.0 & 87.0 & 95.17 & 23.0 \\
D-SFT (ours)             & 8B   & 92.3 & 95.9 & 98.32 & 4.1 & 87.0 & 93.5 & 97.70 & 9.5 \\
\midrule
\multicolumn{10}{l}{\textit{Symbolic RL (D-SFT init., 8B)}} \\
D-SFT{+}DAPO & 8B & 96.2 & 96.2 & 98.64 & 0.2 & 94.3 & 95.7 & 98.67 & 2.0 \\
D-SFT{+}DAPO{+}Short-RL & 8B & 96.9 & 96.9 & 99.23 & 0.0 & 95.2 & 95.5 & 99.32 & 1.5 \\
D-SFT{+}DAPO{+}Short-RL{+}GroupAdapt ($\tau{=}0.5$) & 8B & \textbf{97.3} & \textbf{97.3} & \textbf{99.39} & 0.2 & 94.0 & 95.2 & 99.22 & 2.2 \\
\bottomrule
\end{tabular}
\end{table*}

\subsection{Setup}
\label{sec:setup}

\paragraph{Research questions.}
The experiments evaluate four claims: whether symbolic
rejection-sampling distillation improves compact planners, how close
the distilled planner is to larger open and frontier models, whether
SFT-initialized symbolic-reward RL can improve one-sample
Strict-Pass, and whether Short-RL plus GroupAdapt can reduce length
without sacrificing correctness.

\paragraph{Benchmarks and splits.}
We evaluate on two benchmarks from
BEHAVIOR-1K~\cite{li2023behavior}: \textbf{Behavior-100}
(B-100, 100 tasks, 200 paired single-/dual-arm episodes) and
\textbf{Behavior-1000} (B-1000, 1{,}000 tasks). To better reflect
realistic robot embodiments, each evaluated task is instantiated in
single-arm and dual-arm settings when available. For B-1000 we
reserve a held-out 97-task, 194-episode test set, which also serves
as the RL validation set. The remaining B-1000 tasks form the
training pool and are augmented by LLM-based BDDL perturbation to
2{,}117 training episodes (\Cref{app:augmentation}). For RL
analysis, both B-100 and the B-1000 held-out split are in-domain
embodied evaluations under the same extended-BDDL planning setting.
Following Short-RL~\cite{yuan2025efficient}, we also report
out-of-domain mathematical reasoning on AIME24, AIME25, AMC23, and
MATH500 as a secondary check for whether length control damages
non-embodied reasoning, and it is not an optimization target or a core
claim of this paper.

\paragraph{Metrics.}
We report Strict-Pass (SP: all goals met with zero precondition
errors), Engine-Pass (EP: all goals met while allowing execution
errors), average Goal Completion Ratio (GCR), and average output
length. SP is the primary correctness metric because it requires
both goal satisfaction and executable action sequences. Full
per-embodiment breakdowns are in \Cref{app:arm_comparison}.

\paragraph{Baselines.}
We compare against a one-sample baseline suite spanning frontier API
models (DeepSeek-V4-Flash/Pro, Gemini-3.1-Pro, Kimi-K2.6, GLM-5.1,
GPT-5.4) and a broad set of open-weight models, including
Gemma-4-31B-IT and the Qwen3 / Qwen3.5 / Qwen3.6 families from 8B to
122B active parameters.

\paragraph{Decoding protocol.}
Unless a vendor API forces a different setting, all models are
decoded at temperature~0.6 and top-$p$~0.9 with thinking/reasoning
mode enabled when available. \textbf{Unless stated otherwise,
evaluations use the default \emph{Guided} prompt on the extended-BDDL
scene representation.} For RL checkpoint evaluations, each table entry
and curve point is averaged over three decoding seeds, and non-RL baseline
rows use the one-sample protocol. We discuss prompt sensitivity as a
limitation rather than making prompt engineering a main experimental
axis.

\paragraph{Training and SFT initialization.}
The RL stage in this paper is scoped to a single initialization,
the compact supervised checkpoint \textbf{D-SFT}
(Qwen3-8B-Gemma-Distill-SFT), obtained by symbolic rejection
sampling from Gemma-4-31B-IT. We use the same prompt and symbolic
engine as in evaluation. The SFT-initialized DAPO
configuration uses group size $G{=}8$, and full hyperparameters are in
\Cref{tab:train_hparams}.

\subsection{Main Results: Baseline Comparison}
\label{sec:main_results}

\Cref{tab:main_results} shows a consistent gain from reusing symbolic
verification as training feedback. D-SFT raises Qwen3-8B from 77.3 to
92.3 SP on B-1000 and from 72.0 to 87.0 SP on B-100, and symbolic-reward
DAPO further reaches 96.2 and 94.3 SP. Adding Short-RL keeps this
accuracy while compressing responses, and the default GroupAdapt
configuration reaches the best B-1000 SP and GCR among the 8B variants.
The B-100 result is slightly below Short-RL at the final checkpoint, so
we interpret GroupAdapt as a correctness-preserving compression
strategy rather than a uniform per-dataset accuracy improvement.

\subsection{Ablations and Length Analysis}
\label{sec:ablations}

\begin{figure*}[!t]
    \centering
    \includegraphics[width=\textwidth]{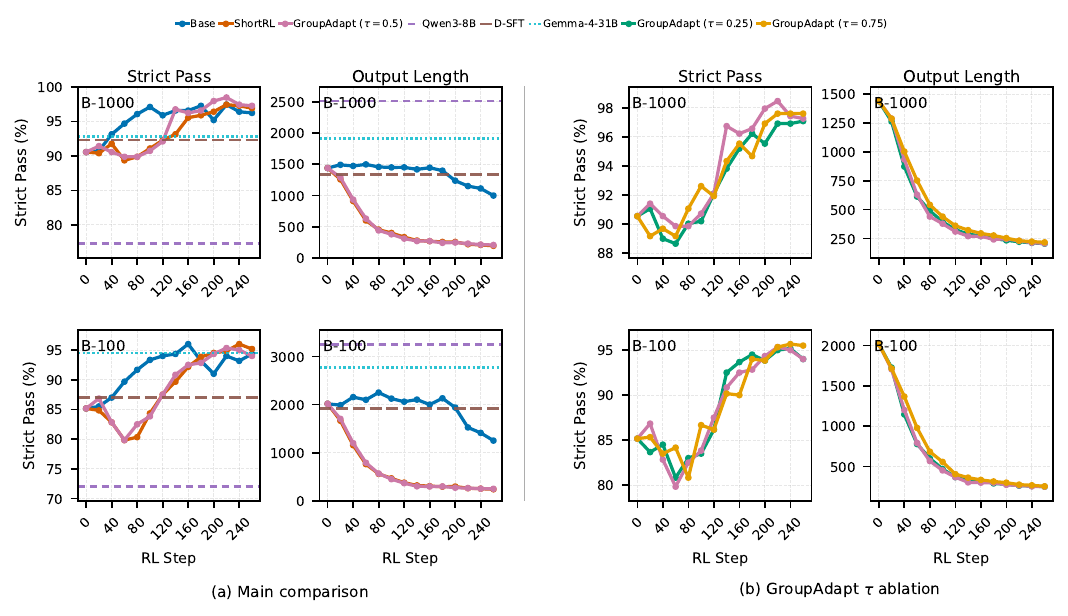}
    \caption{\textbf{Compact symbolic-RL trajectories on in-domain embodied
    validation, averaged over three decoding seeds.} Rows show B-1000
    and B-100. A vertical divider separates (a) the main comparison
    from (b) the GroupAdapt threshold ablation. Each group reports
    Strict-Pass and output length, while GCR and error-rate curves are moved
    to \Cref{fig:appendix_gcr_error_curves}.}
    \label{fig:rl_main_and_tau_curves}
\end{figure*}

\begin{table*}[!t]
\centering
\caption{Ablation axes for correctness and compactness. Rows compare
the no-length-adaptation DAPO baseline, Short-RL, and GroupAdapt
thresholds. GA denotes GroupAdapt. We report in-domain embodied SP,
secondary out-of-domain math accuracy, and final inference length,
averaged over three decoding seeds.}
\label{tab:rl_ablation}
\footnotesize
\renewcommand{\arraystretch}{1.05}
\begin{tabular}{@{}cc cc cccc cc@{}}
\toprule
\textbf{Short-RL} & \textbf{GA} &
\multicolumn{2}{c}{\textbf{ID SP}$\uparrow$} &
\multicolumn{4}{c}{\textbf{Math Acc.}$\uparrow$} &
\multicolumn{2}{c}{\textbf{Len.}$\downarrow$} \\
\cmidrule(lr){3-4} \cmidrule(lr){5-8} \cmidrule(lr){9-10}
& & \textbf{B1K} & \textbf{B100} &
\textbf{A24} & \textbf{A25} & \textbf{AMC} & \textbf{M500} &
\textbf{B1K} & \textbf{B100} \\
\midrule
-- & -- & 96.2 & 94.3 & 37.8 & 24.4 & 72.5 & 73.4 & 999 & 1253 \\
\checkmark & -- & 96.9 & 95.2 & 25.6 & 20.0 & 67.5 & 72.7 & 194 & 242 \\
\checkmark & $\tau{=}0.25$ & 97.1 & 94.0 & 30.0 & 16.7 & 65.8 & 73.1 & 205 & 251 \\
\checkmark & $\tau{=}0.5$ & 97.3 & 94.0 & 30.0 & 21.1 & 65.8 & 73.7 & 207 & 251 \\
\checkmark & $\tau{=}0.75$ & 97.6 & 95.5 & 27.8 & 17.8 & 69.2 & 73.3 & 217 & 255 \\
\bottomrule
\end{tabular}
\renewcommand{\arraystretch}{1.0}
\end{table*}

\Cref{tab:rl_ablation} shows that most of the length reduction comes
from the correctness-gated Short-RL objective: final output length drops
from 999/1253 tokens to 194/242 tokens on B-1000/B-100 while SP is
maintained or improved. GroupAdapt then changes how this pressure is
applied. Larger $\tau$ values mark more rollout groups as hard, giving
still-unstable prompts temporary $400$-token slack before they return to
the standard $200$-token budget after their group pass rate improves.
This explains why $\tau{=}0.75$ yields the strongest final embodied SP
with only a small length increase. We keep $\tau{=}0.5$ as the default
because it is the natural half-pass split selected before the final
checkpoint comparison and gives the most balanced trajectory in
\Cref{fig:rl_main_and_tau_curves}. The math columns are included only
as an OOD side-effect check following Short-RL: length-adapted variants
are not optimized for mathematical derivations, but GroupAdapt recovers
part of Short-RL's math drop on AIME24, AIME25, and MATH500.

\section{Conclusion}
\label{sec:conclusion}

This paper presented \method, a BDDL-centric pipeline for turning
open-world or curated task evidence into verifiable planner training.
By using verified typed objects, initial predicates, and goal
predicates as a shared interface, \method connects data construction,
symbolic verification, SFT filtering, and RL reward design. The
system constructs or accepts BDDL specifications, rewrites them into
hierarchical planning conversations, expands the callable action set
through predicate-guided action discovery, and verifies generated
plans with deterministic GCR, Engine-Pass, Strict-Pass, and error
signals.

Empirically, the pipeline turns the 8B planner into a competitive
symbolic executor while keeping its responses short. The final
GroupAdapt configuration reaches 97.3 SP on B-1000 with an average
response length of 207 tokens, compared with 96.2 SP and 999 tokens
for DAPO without length adaptation. The threshold ablation further
shows that giving low-pass-rate rollout groups temporary slack is a
useful safeguard: harder prompts can first stabilize under a wider
budget and later return to the shorter budget as their group success
rate improves.

The main technical lesson is that verifier reuse simplifies the
entire pipeline. The same symbolic engine that certifies SFT
trajectories also supplies dense RL rewards, separating partial goal
progress, illegal-but-goal-complete plans, and strict executable
success. On top of this reward, Short-RL and GroupAdapt provide a
controlled path from accurate planning to concise planning:
compression is delayed until correctness stabilizes, and the current
group pass rate adapts the length tolerance without extra difficulty
classifiers or simulator calls. The appendix pass@$k$ analysis suggests
that the compact planner still has consistency headroom, motivating
symbolic-reward RL as a way to convert sampled correct behaviors into
more reliable one-sample planning.

\section*{Limitations}
\method is a planning model rather than a low-level control policy.
Its action space is intentionally abstract: for example, a task such
as making coffee is evaluated at the level of symbolic planning steps,
not decomposed into fine-grained manipulation primitives. Learning
such low-level skills would require separate robot-control training.

A second limitation concerns deployment: a real robot must scan a
scene and quickly construct the task-relevant BDDL objects and initial
predicates from perception, while the goal can often be derived from
the user's instruction rather than scanned from the scene. Robust
real-time scene-to-BDDL construction with diverse cameras and
viewpoints therefore remains an open problem.

Finally, our main experiments use the Guided prompt for consistency
with D-SFT training, but forbid-style prompts may be more helpful than
rule-style prompts for stronger models, and prompt choice therefore remains
an evaluation variable rather than a core claim of the method.


\bibliography{custom}

\clearpage
\appendix
\setlength{\floatsep}{6pt plus 1pt minus 1pt}
\setlength{\textfloatsep}{8pt plus 1pt minus 2pt}
\setlength{\intextsep}{6pt plus 1pt minus 1pt}

The appendix follows the same pipeline order as the main paper:
\Cref{app:data} details data construction, \Cref{app:action_builder}
describes the symbolic engine and action library, \Cref{app:hparams}
collects RL training details, and \Cref{app:additional_eval} reports
additional evaluation analyses.

\section{Data Construction Details}
\label{app:data}

This appendix summarizes the conversion from raw BDDL to
planner-facing conversations. Each sample is built by parsing objects,
initial predicates, and goals, identifying the robot type, selecting
the corresponding action library and system prompt, and rewriting the
formal goal into a natural-language request. Distractor objects are
retained to test whether the model identifies task-relevant objects.
This section first describes video-to-BDDL conversion
(\Cref{app:video_to_bddl}), then shows the model-facing sample format
(\Cref{app:data_sample}), and finally summarizes data augmentation
(\Cref{app:augmentation}). \Cref{fig:data_sample} gives one concrete
example.

\subsection{Video-to-BDDL Construction}
\label{app:video_to_bddl}

When the source is a video or other visual scene evidence, \method
uses the task description as a prior for deciding what to look for.
The instruction specifies the target objects and related context
objects that should be grounded, while the video supplies evidence for
where those objects are and what states or relations they currently
satisfy. The conversion has three stages. First, the task description
is parsed into a search plan over relevant object categories, such as
target objects, tools, containers, surfaces, and the agent. Second, the
visual front end follows the Set-of-Mark visual prompting
style~\cite{yang2023set} to select frames that expose these objects
and ground them as BDDL instances, in the spirit of open-vocabulary
detection references such as~\cite{liu2024grounding}, and observable
initial predicates are then created
from spatial and state evidence, including room membership, support,
containment, open/closed state, cleanliness, and similar task-relevant
properties. Third, the task instruction is converted into goal
predicates over the same typed objects. The resulting BDDL is accepted
only after an LLM-as-a-judge
verifier~\cite{zheng2023judging,liu2023g} confirms syntax,
object coverage, initial-state consistency, and goal executability,
matching the four content axes used by the symbolic engine.

\begin{figure}[t]
\centering
\includegraphics[width=\linewidth]{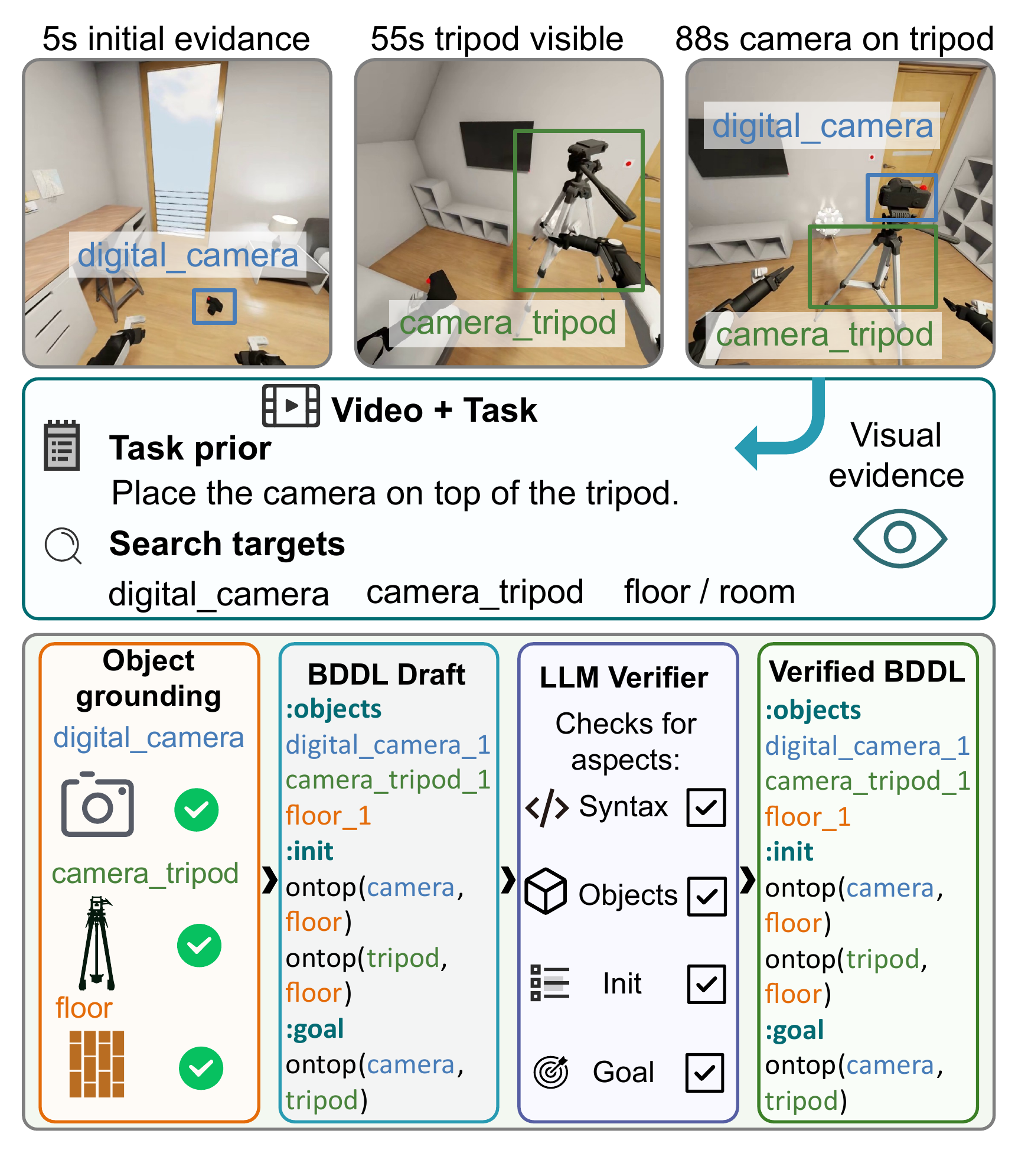}
\caption{\textbf{Task-guided video-to-BDDL construction on a camera episode.}
The task description provides a prior over target and related objects,
while key frames provide visual evidence for grounding these objects
and their initial relations. The BDDL draft is then checked for syntax,
object coverage, initial-state consistency, and goal executability
before being accepted.}
\label{fig:video_to_bddl_pipeline}
\end{figure}

\Cref{fig:video_to_bddl_pipeline} summarizes this construction
pipeline. The important design choice is that the same verified BDDL
serves three roles: it is rewritten into the planner prompt, it defines
the target condition for training examples, and it is reused unchanged
by the symbolic engine when executing generated action code.

\paragraph{Quantitative evaluation.}
Following the pipeline in \Cref{fig:video_to_bddl_pipeline}, we apply
it to 50 BEHAVIOR-1K task videos from our evaluation suite and report
four indicators in \Cref{tab:video2bddl_indicators}.

\begin{table}[t]
\centering
\caption{Video-to-BDDL evaluation on 50 BEHAVIOR-1K task videos.
\emph{Engine Loading}: fraction of drafts the symbolic engine
loads. \emph{Verifier}: fraction the LLM verifier accepts on the
four content axes (\emph{Syntax}, \emph{Objects}, \emph{Init},
\emph{Goal}). \emph{Core Semantic Agreement}: fraction rated
equivalent or partially equivalent to the official BEHAVIOR-1K
BDDL by an independent GPT-5.4 judge across core entities, init,
goal, and overall. \emph{Win Rate}: fraction where the judge
prefers our BDDL over the released BDDL with respect to the
natural-language instruction. Synset and lemma differences (\eg{}
\texttt{basket.n.01} vs.\ \texttt{wicker\_basket.n.01}) and
scene-decoration-only objects are not penalized.}
\label{tab:video2bddl_indicators}
\small
\setlength{\tabcolsep}{4pt}
\begin{tabular}{@{}lc@{}}
\toprule
\textbf{Indicator} & \textbf{Value} \\
\midrule
Engine Loading Rate            & 50/50 (100\%) \\
Verifier Acceptance Rate       & 37/50 (74\%)  \\
Core Semantic Agreement        & 36/50 (72\%)  \\
Instruction Alignment Win Rate & 32/50 (64\%)  \\
\bottomrule
\end{tabular}
\end{table}

The engine and the verifier accept the large majority of drafts
at 100\% and 74\%, with 36/50 drafts agreeing with the released BDDL on core
semantics, and on 32/50 the judge prefers our BDDL with respect to
the natural-language instruction. The 14/50 disagreements are
dominated by scene-decoration objects (sauces, packaging, room
decor) declared in the released BDDL but not required by the
instruction, suggesting that the construction is grounded in the
written task rather than memorising static curation choices.

\begin{figure*}[t]
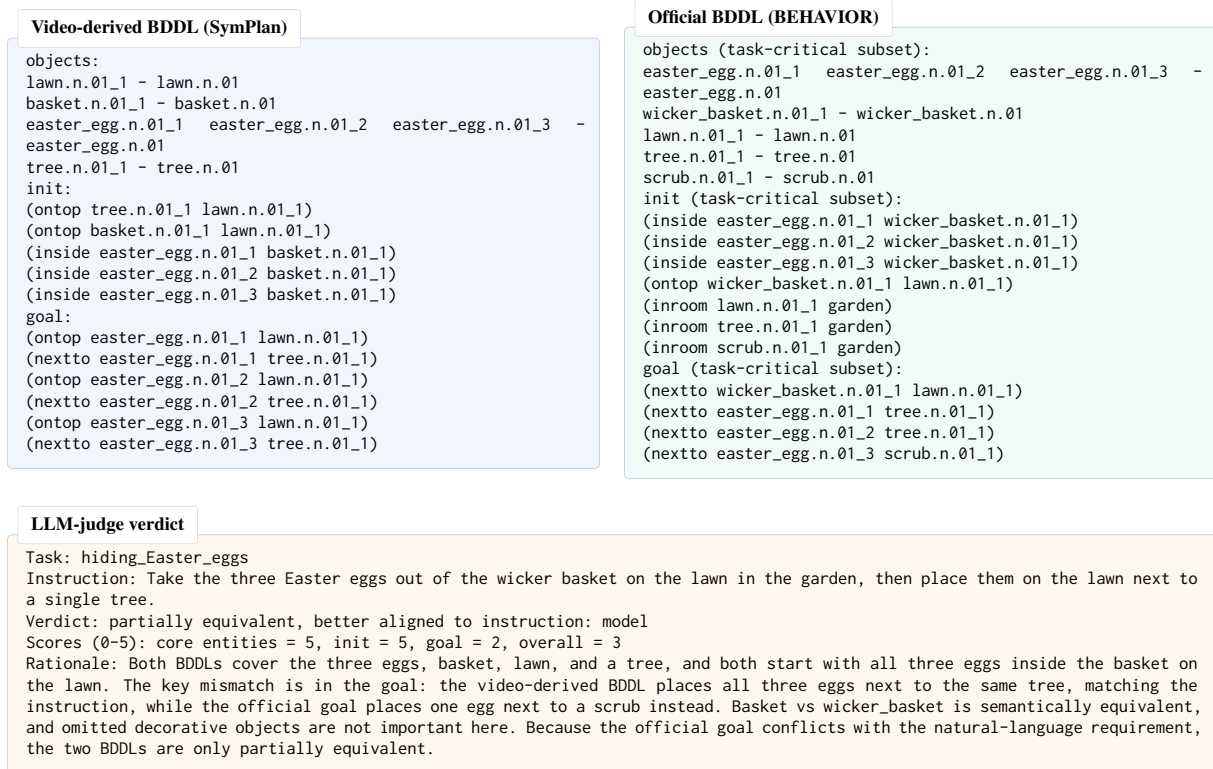

  \centering
  \begin{minipage}{0.49\linewidth}
  \begin{tcolorbox}[samplebox, title=Video-derived BDDL (\method), colback=PromptBlue, fontupper=\scriptsize\ttfamily]
objects:

lawn.n.01\_1 - lawn.n.01

basket.n.01\_1 - basket.n.01

easter\_egg.n.01\_1 easter\_egg.n.01\_2 easter\_egg.n.01\_3 - easter\_egg.n.01

tree.n.01\_1 - tree.n.01

init:

(ontop tree.n.01\_1 lawn.n.01\_1)

(ontop basket.n.01\_1 lawn.n.01\_1)

(inside easter\_egg.n.01\_1 basket.n.01\_1)

(inside easter\_egg.n.01\_2 basket.n.01\_1)

(inside easter\_egg.n.01\_3 basket.n.01\_1)

goal:

(ontop easter\_egg.n.01\_1 lawn.n.01\_1)

(nextto easter\_egg.n.01\_1 tree.n.01\_1)

(ontop easter\_egg.n.01\_2 lawn.n.01\_1)

(nextto easter\_egg.n.01\_2 tree.n.01\_1)

(ontop easter\_egg.n.01\_3 lawn.n.01\_1)

(nextto easter\_egg.n.01\_3 tree.n.01\_1)
\end{tcolorbox}
  \end{minipage}\hfill
  \begin{minipage}{0.49\linewidth}
  \begin{tcolorbox}[samplebox, title=Official BDDL (BEHAVIOR), colback=PromptGreen, fontupper=\scriptsize\ttfamily]
objects (task-critical subset):

easter\_egg.n.01\_1 easter\_egg.n.01\_2 easter\_egg.n.01\_3 - easter\_egg.n.01

wicker\_basket.n.01\_1 - wicker\_basket.n.01

lawn.n.01\_1 - lawn.n.01

tree.n.01\_1 - tree.n.01

scrub.n.01\_1 - scrub.n.01

init (task-critical subset):

(inside easter\_egg.n.01\_1 wicker\_basket.n.01\_1)

(inside easter\_egg.n.01\_2 wicker\_basket.n.01\_1)

(inside easter\_egg.n.01\_3 wicker\_basket.n.01\_1)

(ontop wicker\_basket.n.01\_1 lawn.n.01\_1)

(inroom lawn.n.01\_1 garden)

(inroom tree.n.01\_1 garden)

(inroom scrub.n.01\_1 garden)

goal (task-critical subset):

(nextto wicker\_basket.n.01\_1 lawn.n.01\_1)

(nextto easter\_egg.n.01\_1 tree.n.01\_1)

(nextto easter\_egg.n.01\_2 tree.n.01\_1)

(nextto easter\_egg.n.01\_3 scrub.n.01\_1)
\end{tcolorbox}
  \end{minipage}

  \vspace{0.3em}\begin{tcolorbox}[samplebox, title=LLM-judge verdict, colback=PromptOrange, fontupper=\scriptsize\ttfamily]
Task: hiding\_Easter\_eggs

Instruction: Take the three Easter eggs out of the wicker basket on the lawn in the garden, then place them on the lawn next to a single tree.

Verdict: partially equivalent, better aligned to instruction: model

Scores (0-5): core entities = 5, init = 5, goal = 2, overall = 3

Rationale: Both BDDLs cover the three eggs, basket, lawn, and a tree, and both start with all three eggs inside the basket on the lawn. The key mismatch is in the goal: the video-derived BDDL places all three eggs next to the same tree, matching the instruction, while the official goal places one egg next to a scrub instead. Basket vs wicker\_basket is semantically equivalent, and omitted decorative objects are not important here. Because the official goal conflicts with the natural-language requirement, the two BDDLs are only partially equivalent.
\end{tcolorbox}
  \caption{\textbf{Side-by-side comparison of the video-derived BDDL with the official BEHAVIOR BDDL for the task \texttt{hiding\_Easter\_eggs}.} The LLM judge is instructed to ignore synset/lemma differences (e.g. \texttt{egg.n.02} vs \texttt{easter\_egg.n.01}) and decoration objects, and only score core semantic alignment with the natural-language instruction.}
  \label{fig:appendix_compare_hiding_Easter_eggs}
\end{figure*}

The example in \Cref{fig:appendix_compare_hiding_Easter_eggs}
illustrates a typical disagreement of this kind. Both BDDLs cover the
three Easter eggs, the basket, the lawn, and a tree, and both place
the eggs inside the basket initially. The video-derived goal places
all three eggs next to the same tree, which matches the
natural-language instruction verbatim. The released goal instead
places one egg next to a scrub, which is a curation artifact rather
than a planning requirement, and the judge accordingly counts this case
toward our Instruction Alignment Win Rate. This kind of mismatch
supports our design choice of treating the natural-language task as
authoritative when the visual evidence and the released BDDL
disagree.

\subsection{Data Sample}
\label{app:data_sample}

\begin{figure*}[t]
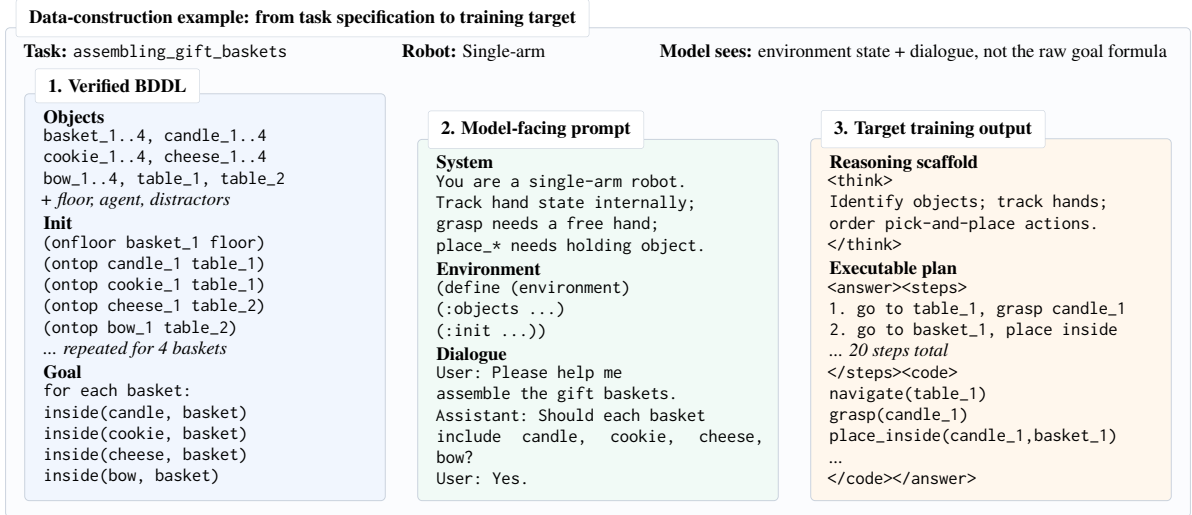

\centering
\begin{tcolorbox}[samplebox,width=0.98\textwidth,colback=PromptGray,title={Data-construction example: from task specification to training target}]
\scriptsize
\textbf{Task:} \texttt{assembling\_gift\_baskets} \hfill
\textbf{Robot:} Single-arm \hfill
\textbf{Model sees:} environment state + dialogue, not the raw goal formula
\vspace{2pt}

\begin{minipage}[t]{0.315\linewidth}
\begin{tcolorbox}[samplebox,colback=PromptBlue,title={1. Verified BDDL}]
\textbf{Objects}\\[-1pt]
\texttt{basket\_1..4, candle\_1..4}\\
\texttt{cookie\_1..4, cheese\_1..4}\\
\texttt{bow\_1..4, table\_1, table\_2}\\
\textit{+ floor, agent, distractors}

\vspace{1pt}
\textbf{Init}\\[-1pt]
\texttt{(onfloor basket\_1 floor)}\\
\texttt{(ontop candle\_1 table\_1)}\\
\texttt{(ontop cookie\_1 table\_1)}\\
\texttt{(ontop cheese\_1 table\_2)}\\
\texttt{(ontop bow\_1 table\_2)}\\
\textit{... repeated for 4 baskets}

\vspace{1pt}
\textbf{Goal}\\[-1pt]
\texttt{for each basket:}\\
\texttt{inside(candle, basket)}\\
\texttt{inside(cookie, basket)}\\
\texttt{inside(cheese, basket)}\\
\texttt{inside(bow, basket)}
\end{tcolorbox}
\end{minipage}
\hfill
\begin{minipage}[t]{0.315\linewidth}
\begin{tcolorbox}[samplebox,colback=PromptGreen,title={2. Model-facing prompt}]
\textbf{System}\\[-1pt]
\texttt{You are a single-arm robot.}\\
\texttt{Track hand state internally;}\\
\texttt{grasp needs a free hand;}\\
\texttt{place\_* needs holding object.}

\vspace{1pt}
\textbf{Environment}\\[-1pt]
\texttt{(define (environment)}\\
\texttt{\quad (:objects ...)}\\
\texttt{\quad (:init ...))}

\vspace{1pt}
\textbf{Dialogue}\\[-1pt]
\texttt{User: Please help me}\\
\texttt{assemble the gift baskets.}\\
\texttt{Assistant: Should each basket}\\
\texttt{include candle, cookie, cheese, bow?}\\
\texttt{User: Yes.}
\end{tcolorbox}
\end{minipage}
\hfill
\begin{minipage}[t]{0.315\linewidth}
\begin{tcolorbox}[samplebox,colback=PromptOrange,title={3. Target training output}]
\textbf{Reasoning scaffold}\\[-1pt]
\texttt{<think>}\\
\texttt{Identify objects; track hands;}\\
\texttt{order pick-and-place actions.}\\
\texttt{</think>}

\vspace{1pt}
\textbf{Executable plan}\\[-1pt]
\texttt{<answer><steps>}\\
\texttt{1. go to table\_1, grasp candle\_1}\\
\texttt{2. go to basket\_1, place inside}\\
\textit{... 20 steps total}\\
\texttt{</steps><code>}\\
\texttt{navigate(table\_1)}\\
\texttt{grasp(candle\_1)}\\
\texttt{place\_inside(candle\_1,basket\_1)}\\
\textit{...}\\
\texttt{</code></answer>}
\end{tcolorbox}
\end{minipage}
\end{tcolorbox}
\caption{\textbf{Illustration of our data construction format.} We parse
the raw BDDL task, including its formal goal, then convert it
into a model-facing prompt consisting of environment state,
robot specification, and multi-turn natural-language dialogue.
The target output follows the training format
\texttt{<think>}\ldots\texttt{</think><answer><steps>}\ldots
\texttt{</steps><code>}\ldots\texttt{</code></answer>}.}
\label{fig:data_sample}
\end{figure*}

\Cref{fig:data_sample} illustrates the conversion from a raw BDDL
task to the model-facing conversation format. The formal BDDL
goal is used during data construction, but it is not exposed
verbatim to the model, and instead, it is rewritten as a natural-language
request in the dialogue.

\subsection{Data Augmentation}
\label{app:augmentation}

We augment the B-1000 training partition with two GPT-4o-based
strategies, summarized in \Cref{tab:aug_strategies}. Each augmented
BDDL is checked for tool consistency, and invalid drafts are excluded
from the training set. Training pairs each task definition with one
robot embodiment, while evaluation uses both single-arm and dual-arm
settings, and the resulting split sizes are reported in \Cref{tab:data_stats}.

\begin{table}[!htbp]
\centering
\caption{Data augmentation strategies and split statistics.}
\begin{subtable}[t]{\linewidth}
\centering
\small
\setlength{\tabcolsep}{6pt}
\begin{tabular}{@{}lll@{}}
\toprule
\textbf{Strategy} & \textbf{What changes} &
  \textbf{Goal changes?} \\
\midrule
Init aug.  & Object placements in \texttt{:init} & No \\
Object aug. & Object types and instances & Yes \\
\bottomrule
\end{tabular}
\caption{Data augmentation strategies.}
\label{tab:aug_strategies}
\end{subtable}

\vspace{0.45em}
\begin{subtable}[t]{\linewidth}
\centering
\scriptsize
\setlength{\tabcolsep}{4pt}
\begin{tabular}{@{}lccccc@{}}
\toprule
\textbf{Split} & \textbf{Original} & \textbf{Init aug.} &
  \textbf{Object aug.} & \textbf{Total defs} & \textbf{Episodes} \\
\midrule
B-1000 train & 959 & 376 & 782 & 2{,}117 & 2{,}117 \\
B-1000 test  & 48  & 9   & 40  & 97      & 194 \\
B-100 test   & 100 & 0   & 0   & 100     & 200 \\
\bottomrule
\end{tabular}
\caption{Training and evaluation data statistics. Augmented BDDL files
are counted as new task definitions.}
\label{tab:data_stats}
\end{subtable}
\end{table}

\section{Symbolic Engine and Action Library}
\label{app:action_builder}

This section documents the execution layer behind the symbolic
feedback used in the main paper. We first replay one verified plan
(\Cref{app:symbolic_replay}), then describe action-set scaling
(\Cref{app:action_scaling}), predicate resolution
(\Cref{app:predicate_resolution}), and the complete action library
(\Cref{app:actions}).

\subsection{Symbolic Engine Replay}
\label{app:symbolic_replay}

\begin{figure}[t]
\centering
\includegraphics[width=\linewidth]{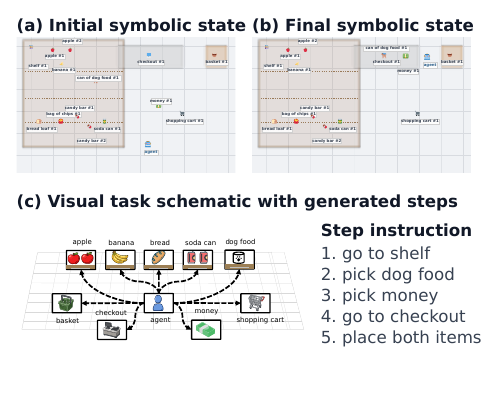}
\caption{\textbf{Symbolic replay for \texttt{buy\_dog\_food}.} The engine
executes generated action code from the initial grocery-store state to
the verified checkout state, using the same BDDL specification that
was used to construct the planner input.}
\label{fig:visual_replay}
\end{figure}

After the planner emits action code, the symbolic engine executes the
code against the same BDDL state and goal. \Cref{fig:visual_replay}
illustrates this replay on a compact grocery-store task. The symbolic
panels show the initial state and the verified final state, while the
visual schematic gives a human-readable view of the task and generated
high-level steps.

\subsection{Action Set Scaling}
\label{app:action_scaling}

\Cref{tab:action_scaling} summarizes the action-set expansion
from the 14-action B-100 engine to the 34-action B-1000 engine.
Predicate gap analysis on B-1000 identifies 23 uncovered predicate
requirements, and after consolidation and LLM review, these become 20
accepted action additions with full resolved-goal coverage.

Concretely, auto-discovery first normalizes B-1000 goal predicates
through alias, regex, composite, and handled-format rules. Predicates
that cannot be mapped to the existing B-100 action effects are counted
as uncovered requirements. The proposal stage groups semantically
equivalent requirements, infers action names and arguments from
predicate morphology and goal co-occurrence patterns, and uses LLM
review to check the proposed action semantics. A proposed action is
accepted only if its emitted effect resolves at least one previously
uncovered requirement without introducing invalid arguments or
conflicting state updates. Coverage is computed after validation as the
fraction of normalized goal-predicate requirements supported by either
the original action set or the accepted additions. The reduction from
23 uncovered requirements to 20 additions is due to consolidation:
several requirements share the same action schema after normalization.

\begin{table}[htbp] 
\centering
\caption{Action-set expansion from B-100 to B-1000. ``Uncovered
goal predicates'' counts post-resolution predicate requirements
not directly supported by the B-100 action set. ``Accepted
additions'' counts the final new actions introduced after
consolidation and LLM review.}
\small
\setlength{\tabcolsep}{6pt}
\begin{tabular}{@{}lcc@{}}
\toprule
& \textbf{B-100} & \textbf{B-1000} \\
\midrule
Base actions           & 14 & 14 \\
Uncovered goal predicates & -- & 23 \\
Reviewed proposals     & -- & 23 \\
Accepted additions     & -- & 20 \\
\midrule
Total actions          & 14 & 34 \\
Resolved goal predicate coverage & 100\% & 100\% \\
\bottomrule
\end{tabular}
\label{tab:action_scaling}
\end{table}

\subsection{Predicate Resolution Rules}
\label{app:predicate_resolution}

The rules used for predicate resolution during the gap analysis are
detailed in \Cref{tab:pred_resolution}.

\begin{table}[!htbp]
\centering
\caption{Predicate resolution and action synthesis rules used in
gap analysis.}
\scriptsize
\setlength{\tabcolsep}{3pt}
\begin{tabularx}{\linewidth}{@{}l>{\raggedright\arraybackslash}X>{\raggedright\arraybackslash}X>{\raggedright\arraybackslash}X@{}}
\toprule
\textbf{Rule} & \textbf{Example} &
\textbf{Resolved} & \textbf{Status} \\
\midrule
Direct    & \texttt{ontop}          & \texttt{ontop}
  & Directly supported \\
Alias     & \texttt{covered\_stain} & \texttt{covered}
  & Alias-resolved \\
Regex     & \texttt{dust.n.01\_1}   & \texttt{dusty}
  & Regex-resolved \\
Composite & \texttt{contains\_peaches} & \texttt{contains}
  & Composite-resolved \\
Handled format & \texttt{inside\_exists} & \texttt{inside\_exists}
  & Handled by goal checker \\
Inverse   & \texttt{toggled\_on = False}  & \texttt{toggle\_off}
  & Accepted addition \\
Missing   & \texttt{folded = True}         & \texttt{fold}
  & Accepted addition \\
\bottomrule
\end{tabularx}
\label{tab:pred_resolution}
\end{table}

\paragraph{Mapping to the implementation.}
The three stages described in \Cref{sec:data_synthesis}
(gap analysis, LLM-assisted proposal, symbolic code synthesis) are
realized by a six-phase implementation: \textsc{scan} and
\textsc{classify} operationalize gap analysis by partitioning the
observed goal predicates into covered / format-variant / truly
missing buckets, while \textsc{infer} and \textsc{propose} operationalize
LLM-assisted proposal by combining morphology-based parameter /
verb inference, goal co-occurrence mining, and optional LLM
review, and \textsc{validate} and \textsc{emit} operationalize
symbolic code synthesis by scoring the proposed set against a
reference library and emitting engine code, tool prompts, and
allowed-function patches. This factorization is an engineering
detail and is orthogonal to the paper's claims, and it only affects
the ease of extending the action library to new predicate
families.

\subsection{Complete Action List}
\label{app:actions}

\Cref{tab:base_actions,tab:b1000_actions} present the complete action
library, separating the base B-100 actions from the B-1000 additions.

\begin{table}[t]
\centering
\caption{Base action library used for B-100.}
\label{tab:base_actions}
\scriptsize
\setlength{\tabcolsep}{2.5pt}
\renewcommand{\arraystretch}{0.92}
\begin{tabularx}{\linewidth}{@{}ll>{\raggedright\arraybackslash}X@{}}
\toprule
\textbf{Action} & \textbf{Args} & \textbf{Effect} \\
\midrule
navigate & obj & close\_to target \\
grasp & obj & inside agent; clear ontop \\
place\_on\_top & obj,tgt & ontop target \\
place\_inside & obj,tgt & inside target \\
place\_next\_to & obj,tgt & nextto target \\
place\_under & obj,tgt & under target \\
open & obj & open = True \\
close & obj & open = False \\
toggle\_on & obj & toggled\_on = True \\
cut & obj & sliced = True \\
pour & obj,tgt & covered target \\
clean & obj & stained/dusty = False \\
wait\_for\_cooked & time & cooked = True \\
soak & obj,tgt & soaked = True \\
\bottomrule
\end{tabularx}
\renewcommand{\arraystretch}{1.0}
\end{table}

\begin{table}[t]
\centering
\caption{Actions added for B-1000 after predicate-gap analysis and LLM
review.}
\label{tab:b1000_actions}
\scriptsize
\setlength{\tabcolsep}{2.5pt}
\renewcommand{\arraystretch}{0.92}
\begin{tabularx}{\linewidth}{@{}ll>{\raggedright\arraybackslash}X@{}}
\toprule
\textbf{Action} & \textbf{Args} & \textbf{Effect} \\
\midrule
toggle\_off & obj & on = False; toggled\_on = False \\
fill & obj,tgt & filled target \\
fold & obj & folded = True \\
unfold & obj & unfolded = True; folded = False \\
attach & obj,tgt & attached target; attached\_to target \\
screw & obj,tgt & screwed target \\
overlay & obj,tgt & overlaid target \\
drape & obj,tgt & draped target \\
heat & obj & hot = True \\
water & obj & watered = True; wet = True; dry = False \\
saturate & obj,tgt & saturated target \\
paint & obj & painted = True \\
set\_timer & obj & timeset = True \\
make & obj & real = True \\
repair & obj & broken = False \\
break\_obj & obj & broken = True \\
burn & obj & burnt = True \\
ignite & obj & on\_fire = True \\
patch & obj & torn = False; patched = True \\
uncrimp & obj & crumpled = False \\
\bottomrule
\end{tabularx}
\renewcommand{\arraystretch}{1.0}
\end{table}

\section{Training and RL Details}
\label{app:hparams}

This section contains the RL configuration and reward details that
support \Cref{sec:reward_design,sec:length_compression}. We list the
shared DAPO hyperparameters in \Cref{tab:train_hparams} and the reward
ordering in \Cref{app:reward_landscape}.

\begin{table*}[!t]
\centering
\caption{Core hyperparameters for SFT-initialized DAPO. All RL
variants share this configuration and differ only in the length-reward
term and the GroupAdapt tolerance schedule.}
\label{tab:train_hparams}
\scriptsize
\setlength{\tabcolsep}{3pt}
\renewcommand{\arraystretch}{0.92}
\begin{tabularx}{\textwidth}{@{}l>{\raggedright\arraybackslash}Xl>{\raggedright\arraybackslash}X@{}}
\toprule
\textbf{Hyperparameter} & \textbf{Value} &
\textbf{Hyperparameter} & \textbf{Value} \\
\midrule
Initialization checkpoint & D-SFT (Qwen3-8B-Gemma-Distill-SFT) &
RL algorithm & DAPO~\cite{yu2026dapo} \\
Advantage estimator & group-relative (GRPO-style) &
Group size $G$ & 8 \\
Train batch size & 32 &
Generation batch size & 96 \\
Training sampling & temp $0.8$, top-$p$ $0.9$ &
Validation sampling & temp $0.6$, top-$p$ $0.9$ \\
Learning rate & $1\times 10^{-6}$ &
LR warmup steps & 10 \\
Weight decay & 0.1 &
Gradient clip & 1.0 \\
PPO clip (low / high) & $0.2$ / $0.28$ (Clip-Higher) &
Dual-clip ratio $c$ & 10.0 \\
Loss aggregation & token-mean &
Dynamic sampling & enabled, metric = total reward \\
Max regeneration batches & 50 &
Overlong shaping & buffer $=1048$, factor $=1.0$ \\
Max prompt length & 6144 &
Max response length & 8192 \\
PPO mini-batch size & 32 &
PPO micro-batch / GPU & 4 \\
\bottomrule
\end{tabularx}
\renewcommand{\arraystretch}{1.0}
\end{table*}

\subsection{Reward Landscape Details}
\label{app:reward_landscape}

\Cref{tab:reward_landscape} lists the reward values produced by
the multi-granular symbolic reward (\Cref{sec:reward_design}) for
representative plan outcomes. The ordering ensures that complete,
error-free plans (Strict Pass) always receive the highest reward,
while partial successes are rewarded proportionally.

\begin{table}[!htbp]
\centering
\caption{Reward landscape for representative symbolic outcomes.}
\label{tab:reward_landscape}
\scriptsize
\setlength{\tabcolsep}{4pt}
\renewcommand{\arraystretch}{0.92}
\begin{tabular}{@{}lcccr@{}}
\toprule
\textbf{Outcome} & GCR & Err? & EP? & $R_{\text{ans}}$ \\
\midrule
Strict Pass     & 1.0 & No  & Yes & \phantom{$-$}2.5 \\
EP-only         & 1.0 & Yes & Yes & \phantom{$-$}1.5 \\
Near-miss (0.8) & 0.8 & No  & No  & \phantom{$-$}1.5 \\
Near-miss (0.6) & 0.6 & No  & No  & \phantom{$-$}1.0 \\
Partial         & 0.5 & Yes & No  & $-$0.25 \\
Failure         & 0.0 & No  & No  & $-$0.5 \\
\bottomrule
\end{tabular}
\renewcommand{\arraystretch}{1.0}
\end{table}

\section{Additional Evaluation Analysis}
\label{app:additional_eval}

This section collects analyses that support but are not required for
the main comparison: additional RL curves (\Cref{app:additional_rl_curves}),
embodiment-specific behavior (\Cref{app:arm_comparison}), task-horizon
statistics (\Cref{app:task_complexity}), and pass@$k$ headroom
(\Cref{app:pass_at_k}).

\subsection{Additional RL Curves}
\label{app:additional_rl_curves}
\begin{figure*}[tbp]
    \centering
    \includegraphics[width=\textwidth]{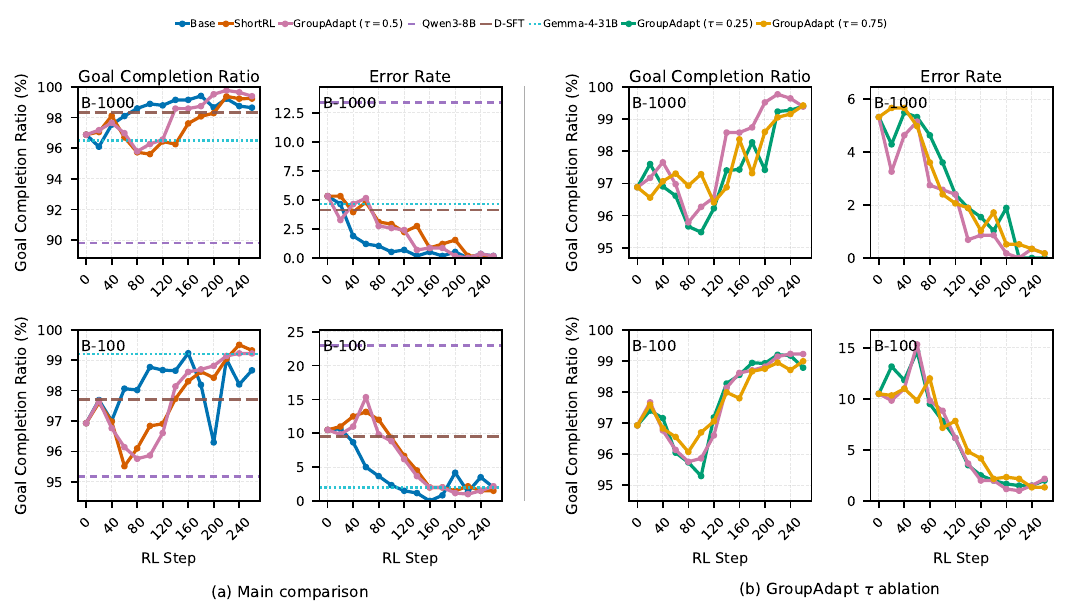}
    \caption{\textbf{Additional symbolic-RL trajectories, averaged over three
    decoding seeds.} The layout matches \Cref{fig:rl_main_and_tau_curves},
    with a vertical divider separating (a) the main comparison from (b) the
    GroupAdapt threshold ablation. Each group reports Goal Completion
    Ratio and engine error rate.}
    \label{fig:appendix_gcr_error_curves}
\end{figure*}

\Cref{fig:appendix_gcr_error_curves} complements the main SP/length
curves by reporting Goal Completion Ratio and engine error rate. GCR
shows whether methods preserve partial task progress, while the
error-rate curves show whether higher SP comes from cleaner executable
plans rather than merely satisfying goals with invalid intermediate
actions.

\subsection{Single-arm vs.\ Dual-arm Comparison}
\label{app:arm_comparison}

\Cref{tab:arm_diff} summarizes the single-arm and dual-arm setup,
and \Cref{tab:arm_perf_b1000} reports the comparison on B-1000.

\begin{table}[!htbp]
\centering
\caption{Key differences between single-arm and dual-arm
configurations. The average command count is computed on B-1000
across the baseline suite in \Cref{tab:arm_perf_b1000}.}
\scriptsize
\setlength{\tabcolsep}{3pt}
\begin{tabularx}{\linewidth}{@{}l>{\raggedright\arraybackslash}X>{\raggedright\arraybackslash}X@{}}
\toprule
\textbf{Aspect} & \textbf{Single-arm} & \textbf{Dual-arm} \\
\midrule
Max held objects & 1 & 2 \\
Action interface & \multicolumn{2}{c}{Same (\eg
  \texttt{grasp(obj)}, \texttt{place\_*(obj, tgt)})} \\
Key constraint   & Must release before next grasp &
  Can carry two objects simultaneously \\
System prompt    & ``single-arm robot'' &
  ``dual-arm robot'' \\
Avg.\ commands & 19.6 & 16.0 \\
\bottomrule
\end{tabularx}
\label{tab:arm_diff}
\end{table}

On B-1000, dual-arm settings usually reduce executable command count,
while changes in error rate and goal completion remain more
model-dependent. This suggests that embodiment mainly affects
coordination burden and plan efficiency rather than the semantic
difficulty of the task itself.

\begin{table}[t]
\centering
\caption{Single-arm vs. dual-arm performance on B-1000. Each entry
reports single-arm / dual-arm values. Goal completion is reported as a
fraction, and error rate is reported in percent.}
\label{tab:arm_perf_b1000}
\scriptsize
\setlength{\tabcolsep}{2.5pt}
\renewcommand{\arraystretch}{0.95}
\resizebox{\linewidth}{!}{%
\begin{tabular}{@{}lccc@{}}
\toprule
\textbf{Model} & \textbf{Goal completion} & \textbf{Error rate (\%)} & \textbf{Commands} \\
\midrule
DeepSeek-V4-Flash & 0.987 / 0.980 & 0.03 / 0.09 & 19.09 / 16.03 \\
DeepSeek-V4-Pro & 0.978 / 0.961 & 1.11 / 0.71 & 27.20 / 21.95 \\
Gemini-3.1-Pro & 0.897 / 0.890 & 0.00 / 0.02 & 17.77 / 14.73 \\
Kimi-K2.6 & 0.958 / 0.979 & 0.28 / 0.11 & 13.28 / 10.03 \\
GLM-5.1 & 0.966 / 0.971 & 0.06 / 0.26 & 17.98 / 17.78 \\
GPT-5.4 & 0.964 / 0.943 & 0.00 / 0.07 & 19.96 / 17.18 \\
Qwen3.5-122B & 0.939 / 0.937 & 0.22 / 0.38 & 19.11 / 15.39 \\
Gemma-4-31B & 0.959 / 0.971 & 0.07 / 0.12 & 17.47 / 14.97 \\
Qwen3.6-35B & 0.975 / 0.929 & 0.40 / 0.20 & 18.60 / 13.43 \\
Qwen3.5-35B & 0.975 / 0.887 & 1.05 / 2.44 & 30.91 / 24.82 \\
Qwen3.6-27B & 0.973 / 0.972 & 0.08 / 0.09 & 17.05 / 14.52 \\
Qwen3.5-27B & 0.955 / 0.919 & 0.22 / 0.31 & 18.97 / 15.25 \\
Qwen3-8B & 0.920 / 0.877 & 0.86 / 0.46 & 17.99 / 14.69 \\
D-SFT & 0.979 / 0.988 & 0.06 / 0.13 & 18.40 / 13.81 \\
\bottomrule
\end{tabular}%
}
\renewcommand{\arraystretch}{1.0}
\end{table}

\subsection{Task Complexity Analysis}
\label{app:task_complexity}

To characterize benchmark difficulty, we analyze the distribution
of \emph{executable command count} (the number of primitive robot
actions in each generated plan) on \textbf{B-100} and
\textbf{B-1000}. This metric reflects task
complexity: more commands correspond to multi-step tasks with
more objects or longer action sequences. Command count is distinct
from response-token length: it measures the symbolic action horizon
rather than verbal verbosity.

\begin{figure}[!htbp]
\centering
\includegraphics[width=\linewidth]{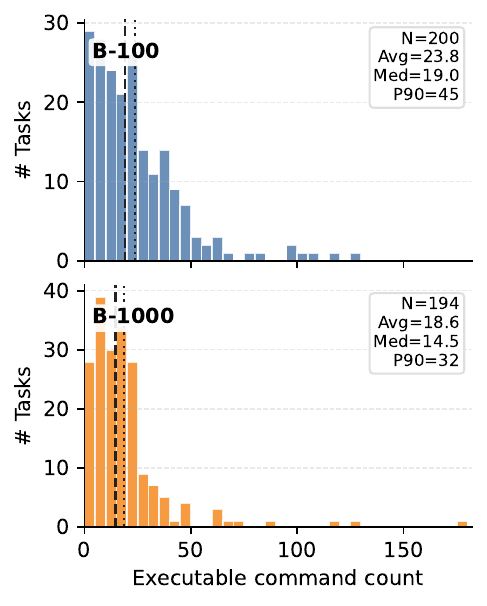}
\caption{\textbf{Histogram of executable command count for GPT-5.4 on B-100
(top, $N$\,=\,200) and B-1000 (bottom, $N$\,=\,194).} Dashed lines mark
the median and dotted lines mark the average, and each panel also reports
$N$, average, median, and P90.}
\label{fig:cmd_count_gpt54}
\end{figure}

\Cref{fig:cmd_count_gpt54} summarizes the executable command
count for GPT-5.4 with per-bin histograms and summary statistics. The
model produces a median of 19 commands on B-100 (average 23.8,
P90 = 45) and a median of 14.5 commands on B-1000 (average 18.6,
P90 = 32), showing a heavier B-100 action-horizon distribution. The
right-skewed tails indicate that some tasks require substantially
longer symbolic action sequences than the median case.

\subsection{Pass@$k$ Analysis: RL Headroom}
\label{app:pass_at_k}

To assess how much reinforcement learning can still improve the
distilled checkpoint D-SFT, we perform a pass@$k$ analysis on both
B-100 and B-1000 using the guided extended-BDDL multi-sample
evaluation. For each task occurrence we draw $n{=}10$ independent
samples with temperature $0.6$ and $\text{top-}p{=}0.9$, and evaluate
each with the symbolic engine. We then compute pass@$k$ using the
unbiased estimator in \Cref{eq:pass_at_k} from
Chen~\etal~\cite{chen2021evaluating}:
\begin{equation}
  \text{pass@}k = \mathbb{E}_{\text{tasks}}\!\left[
    1 - \frac{\binom{n-c}{k}}{\binom{n}{k}}
  \right]
  \label{eq:pass_at_k}
\end{equation}
where $n$ is the total number of samples per task occurrence and $c$
is the number of strict-passing samples.

\Cref{tab:pass_at_k} shows that D-SFT has non-trivial consistency
headroom even after distillation. Strict pass@1 is already high, but
pass@10 rises to 97.0\% on B-100 and 99.0\% on B-1000, leaving
11.6 and 7.6 points of recoverable one-sample reliability,
respectively. This is the kind of gap that group-based symbolic RL is
designed to close.

\begin{table}[htbp]
\centering
\caption{Strict-Pass pass@$k$ for D-SFT on guided extended-BDDL
B-100 and B-1000. $n{=}10$ independent samples are drawn per task
occurrence with temperature $0.6$ and $\text{top-}p{=}0.9$.
$\Delta_{\text{SP}}$ denotes the gap from pass@1.}
\label{tab:pass_at_k}
\small
\setlength{\tabcolsep}{6pt}
\begin{tabular}{@{}lcccc@{}}
\toprule
\multicolumn{1}{c}{}
    & \multicolumn{2}{c}{\textbf{B-100} (200 pairs)}
    & \multicolumn{2}{c}{\textbf{B-1000} (194 pairs)} \\
\cmidrule(lr){2-3} \cmidrule(lr){4-5}
$k$ & SP & $\Delta_{\text{SP}}$ & SP & $\Delta_{\text{SP}}$ \\
\midrule
1   & 85.4 & --    & 91.4 & --    \\
2   & 91.7 & +6.3  & 96.1 & +4.7  \\
3   & 93.6 & +8.2  & 97.8 & +6.5  \\
5   & 95.3 & +9.9  & 98.8 & +7.4  \\
10  & 97.0 & +11.6 & 99.0 & +7.6  \\
\bottomrule
\end{tabular}
\end{table}

\paragraph{Consistency analysis.}
\Cref{tab:consistency} breaks down how reliably D-SFT solves each
task occurrence across the same $n{=}10$ samples. B-100 has a larger
``sometimes pass'' bucket than B-1000, suggesting that the smaller but
structurally richer split still contains more prompts where the policy
can solve the task but does not do so consistently.

\begin{table}[htbp]
\centering
\caption{Task consistency breakdown for D-SFT ($n{=}10$ samples per
task occurrence). ``Always/Sometimes/Never pass'' partition task
occurrences by the number of strict-passing samples out of 10.}
\label{tab:consistency}
\small
\setlength{\tabcolsep}{5pt}
\begin{tabular}{@{}lcc@{}}
\toprule
\textbf{Category} & B-100 & B-1000 \\
\midrule
Always pass (all 10)    & 61.0\% & 76.3\% \\
Sometimes pass (1--9)   & 36.0\% & 22.7\% \\
Never pass (0/10)       & 3.0\%  & 1.0\% \\
\bottomrule
\end{tabular}
\end{table}

\paragraph{Length--correctness relationship.}
Across the B-1000 multi-sample run, failing responses remain on
average noticeably longer than strict-passing ones for the same
prompt. We do not interpret this gap as evidence that verbosity
directly causes errors: harder tasks naturally induce both longer
outputs and more failures. Rather, we take it as a practical
signal that length cannot be reduced uniformly across samples:
trimming long correct plans is unlikely to help, while trimming
long failing plans may amount to truncating genuinely needed
reasoning. This observation directly motivates the
correctness-gated compression used in the final RL stage.

\paragraph{GroupAdapt threshold.}
The same consistency view also motivates the group-adaptive tolerance
in \Cref{eq:adaptive_tolerance}. During DAPO, each prompt is
represented by a rollout group of size $G{=}8$. We therefore treat
groups with at least half of the rollouts passing as \emph{easy}
(4--8 / 8), and the remaining groups as \emph{hard} (0--3 / 8).
Reliable prompts can be compressed under the Short-RL budget, while
less reliable prompts keep wider length slack until their pass rate
improves. The resulting two-level schedule keeps
$\delta_{\text{base}}{=}200$ on easy groups and widens to $400$ tokens
on hard groups (\Cref{eq:adaptive_tolerance}). The main configuration
uses $\tau{=}0.5$, and the ablation in \Cref{fig:rl_main_and_tau_curves}
additionally evaluates $\tau\in\{0.25,0.75\}$.


\end{document}